\documentclass{article} % For LaTeX2e
\usepackage{arxiv,times}

\usepackage{amsmath,amsfonts,bm}

\def\eqref#1{equation~\ref{#1}}
\def\1{\bm{1}}

\DeclareMathAlphabet{\mathsfit}{\encodingdefault}{\sfdefault}{m}{sl}
\SetMathAlphabet{\mathsfit}{bold}{\encodingdefault}{\sfdefault}{bx}{n}

\usepackage{hyperref}
\usepackage[capitalize]{cleveref}
\usepackage{overpic}
\usepackage{xcolor}
\usepackage{url}
\usepackage{graphicx}
\usepackage{amsmath}
\usepackage{algorithm}
\usepackage{algorithmic}
\usepackage{fontawesome}

\title{
Tracking and triangulating firefly flashes \\ in field recordings: 
\texttt{firefl$eye$-net + oorb}
}
\author{Rapha\"el Sarfati \\
Cornell University\\
\texttt{raphael.sarfati@cornell.edu}
}
\iclrfinalcopy

\begin{document}
\maketitle

\begin{abstract}
%Properly identifying firefly flashes in images and movie frames is a central aspect of modern and tracking techniques of Lampyrids.
%While an easy task in perfectly dark conditions, it becomes difficult in the presence of various artifacts, such as crepuscular conditions or anthropogenic light pollution.
Identifying firefly flashes from other bright features in nature images is complicated.
I provide a training dataset and trained neural networks for reliable flash classification.
The training set consists of thousands of cropped images (patches) extracted by manual labeling from video recordings of fireflies in their natural habitat.
The trained network appears as considerably more reliable to differentiate flashes from other sources of light compared to traditional methods relying solely on intensity thresholding.
This robust tracking enables a new calibration-free method for the 3D reconstruction of flash occurrences from stereoscopic 360-degree videos, which I also present here.

\end{abstract}

\section{Introduction}
\begin{figure}
\centering
\includegraphics[width=\linewidth]{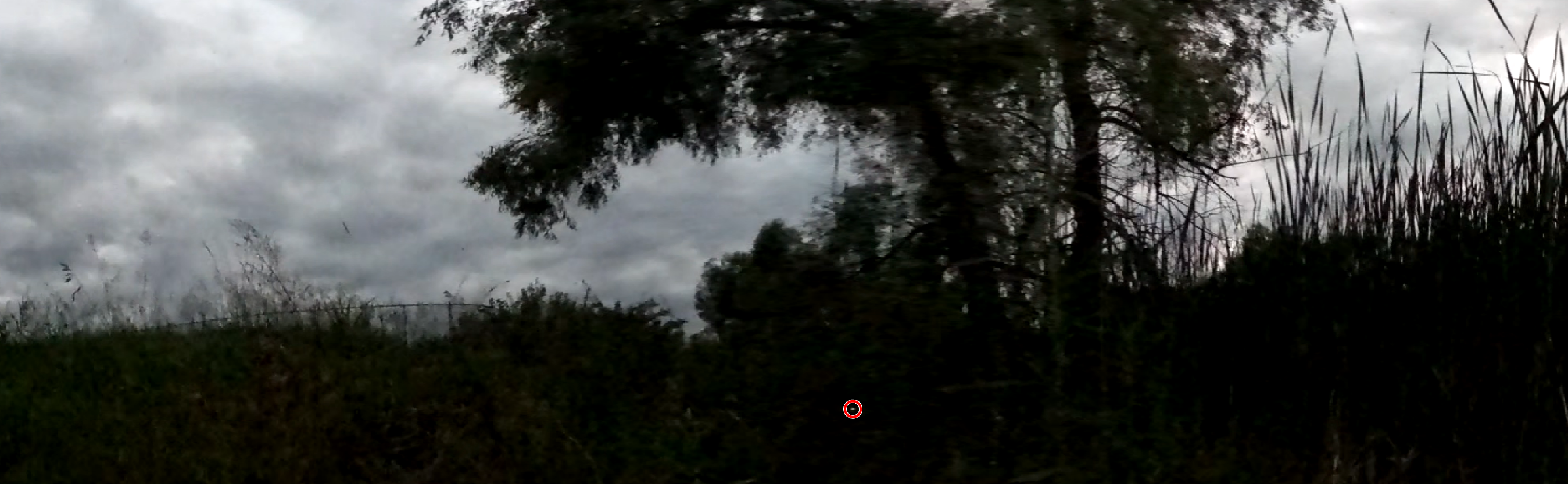}
\caption{
\label{fig:field}
In suboptimal recording conditions, there is a lot of bright artifacts in images.
Intensity thresholding methods usually fail to isolate only flashes (inside red circles) from other bright spots, especially between tree leaves.
A trained \texttt|net|, however, succeeds.
}
\end{figure}

%\texttt{firefl$^e$y$^e$} \\ 
%\texttt{firefl\raisebox{1ex}{\small e}y\raisebox{1ex}{\small e}} \\
%\texttt{firefl\raisebox{1ex}{\tiny e}y\raisebox{1ex}{\tiny e}} \\
Reliable identification of firefly flashes in natural images is either trivial or convoluted.
In the ideal recording conditions of a perfectly dark environment, firefly flashes are simply transient blobs of bright pixels on a dark background. 
A simple intensity thresholding operation is generally sufficient.
In many situations, however, recordings in the wild contain different types of interference or artifacts, which can make intensity thresholding largely inefficient (Figure~\ref{fig:field}). 
These artifacts include:
(A1) significant ambient light, for crepuscular species or recordings in places with substantial light pollution (e.g., near urban areas); (A2) persistent light pollution, such as the moon or distant backyard or street lights; (A3) transient light pollution, such as car headlights, lightning, or a nocturnal hiker's flashlight.

Here, I present a Deep Learning approach which achieves far superior performance compared to traditional feature engineering methods.
I trained a Convolutional Neural Network (CNN) on manually labeled patches extracted from movie frames, and use the trained network to classify bright blobs as flash or artifact.
The \verb|firefl|\faEye~dataset and trained networks (\verb|net|) are provided on the project's repository: \href{https://github.com/rapsar/firefl-eye-net}{github.com/rapsar/firefl-eye-net}.

In previous work, tracking firefly flashes in stereoscopic pairs of videos permitted the 3D reconstruction of blinking swarms~\citep{Sarfati2020}.
However, it required an in-field calibration procedure, which was cumbersome and prone to errors.
With this improved \verb|firefleye-net| tracking, I show that calibration can be performed directly from the data.
I briefly describe the calibration-free algorithm, \verb|oorb|, and show example of 3D reconstruction (\href{https://github.com/rapsar/oorb.git}{github.com/rapsar/oorb}). 

% Fireflies enjoy a rare status among insects: they are (almost) universally appreciated.
% Not only do they not bite, sting, or ooze; many species in this subfamily of beetles (Coleoptera) produce bright but brief flashes during summer nights, turning open fields and deep forests into a poetic light show.
% In many regions across the globe, these bioluminescent displays are associated with fond memories of summertime evenings.
% Thanks to their special charisma, fireflies are often an easy entryway into entomology and insect conservation~\cite{Faust2004}.

% Unfortunately, like most biodiversity, firefly populations globally tend to decline, due notably to habitat loss, increased light pollution, and climate change~\cite{Fallon2022}.
% Careful monitoring of population strength, coupled with accurate species identification from flash patterns and insect morphology, should ensure that protect this important natural resources. 

% Traditional firefly research has mostly relied on patient observations and note-taking by passionate human observers~\cite{Faust2017}.
% In recent years, however, modern imaging techniques and computer vision have enabled more quantitative approaches.
% They notably allowed the collection of large datasets which provide rigorous information about flash duration, intensity pattern, flash sequences, spatial trajectories, and density estimates~\cite{Sarfati2020}.

\section{Flash identification with \texttt{firefl}\faEye\texttt{-net}}

\subsection{Initial approach to flash identification}

Previously, I have used adaptive background subtraction (ABS) to correct for artifacts (A1) and (A2). 
By creating a moving average frame of the preceding few seconds, it is possible to account for a bright background and persistent luminous objects when looking for flashes.
Transient light pollution (A3) was corrected in post-processing by discarding frames with an anomalous number of flashes. 
This would sometimes result in discarding a substantial fraction of recordings.

The problem with ABS, however, is that it is sensitive to motion between frames. 
If, for example, the wind is moving the vegetation in front of a radiant sky, the motion will be wrongly identified as possible flashes.

\subsection{The \texttt{firefl-eye} training set}
%In order to improve firefly flash tracking in movies recorded in various conditions, I turned to a machine learning approach. 
Using recordings of firefly displays in natural habitats of various conditions, I build a training set by manual labeling. 
I then trained a small CNN on this dataset to classify bright blobs as a flash or not.

\begin{figure}
\centering
\includegraphics[width=\linewidth]{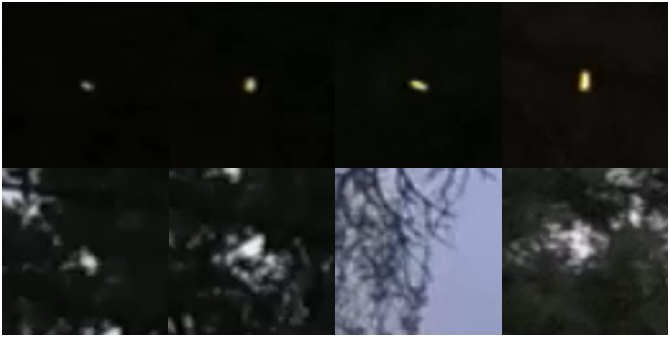}
\caption{
\label{fig:patches}
Example of $65 \times 65$~pixels$^2$ RGB patches. Top row: flashes. Bottom row: bright artifacts.
}
\end{figure}

I manually labeled hundreds of firefly flashes from movie frames, creating a set of patches containing a flash at the center (Fig.~\ref{fig:patches}).
A complementary set of flashes not containing firefly flashes was also created.

The \verb|firefl-eye| training set consists of $65\times65$~pixels$^2$ patches in RGB.
Pixel values range from 0 to 255 (unsigned integers).
The brightest pixel stands at the center (coordinates: 33,33).
These rather large patches allow for a significant amount of contextual information around a firefly flashes.
Most notably, if the bright spot is a foliage artifact, the patch will contain a number of variations around the central pixel (Fig.~\ref{fig:patches}).

The RGB coordinates potentially allow to capture some features relative to the emission spectrum of firefly flashes, which typically peaks in the yellow-green wavelengths (G channel).
It is however unclear how much of these variations is captured by inexpensive cameras and whether they may have an impact on flash classification.

These patches were collected by manually loading recordings of fireflies in the natural habitat and clicking on flash instances.
Flash are rather easily identified by eye, notably thanks to their transient character.
For data consistency, the brightest pixel in the vicinity of the click is then adjusted to be at the center of the patch.

% I then trained a convolutional neural network (CNN) to accurately classify firefly flashes after pre-identification by ABS.

% This approach has proved much more reliable to accurately identify firefly flashes, even in situations with complex artifacts. 
% Of note, I incorporated a trained network into new software \verb|oorbit|, which performs the automated calibration and subsequent 3D reconstruction of firefly swarms from stereoscopic footage from 360-degree.

% The training set is made available at \href{https://github.com/rapsar/FireflEyeNet}{github.com/rapsar/firefl-eye-net}, along with trained networks.

\subsection{The \texttt{net} neural network}
%\subsubsection{Recurrent Neural Network}

The \verb|firefl-eye| dataset was used to train a CNN that would properly classify the bright spot at a center of a patch as either a firefly flash or an artifact.
The network architecture is available in the Github.
Briefly, it consists of a stack of convolutional layers, with activation layers in between.
Various architectures have been explored, with similar performance.

%The neural network architecture is outlined in Figure XXX.
% Briefly, the input layer is XXXX
% The output (classifier) layer follows from a softmax classification to determine whether a detected blob is a firefly flash or not.

% In between, a series of XXX layers 

% \subsubsection{Convolutional Neuronal Network}
% Using a Convolutional Neural Network (CNN) would allow to directly identify flashes in a large movie frame without resorting to pre-processing by intensity thresholding. 
% In practice, however, it seems like this approach is much less efficient.
% There may be several reasons to explain

\subsection{Validation}
\begin{figure}
\centering
\includegraphics[width=\linewidth]{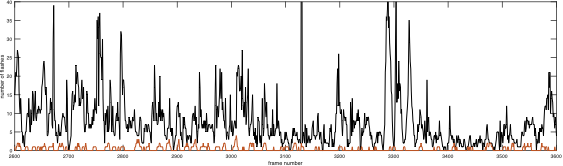}
\caption{
\label{fig:validation}
Comparison of flash tracking with and without \texttt{net} in a noisy recording.
Without \texttt{net} classification, even with ABS, the tracker picks up plenty of false positives (black line), here especially because of wind moving vegetation.
In contrast, by applying net, only true flashes are conserved (orange line).
}
\end{figure}

The neural network approach is found to be much more accurate and reliable for properly identifying firefly flashes in natural images, even in the presence of substantial ambient luminosity, persistent and transient light pollution, and background motion.
Figure~\ref{fig:validation}, for example, shows an example of a video where the wind shakes branches in the field-of-view, which creates a very large number of false positives using ABS.
However, if ABS is first applied to extract possible flashes, which are then classified with \verb|ffnet|, only true flashes remain (orange trace).
%Therefore, it is now incorporated in prior tracking software that has been used for flash identification and firefly tracking.

\section{Calibration-free 3D reconstruction with \texttt{oorb}}

\subsection{3D calibration from stereo recordings}
3D reconstruction relies on the complementary recording of the same scene from different points-of-view.
Corresponding points in both frames are triangulated based on epipolar geometry.
This require, notably, that cameras be calibrated both temporally and spatially.
Temporal calibration means aligning the two frame streams.
Spatial calibration means estimating the relative camera pose (translation + rotation).

\subsection{3D reconstruction with 360-degree cameras}
In recent year, we extended the technique to 360-degree cameras to perform the reconstruction of firefly flashes in 3D while recording from within~\citep{Sarfati2020}.
The technique proved successful and useful; however, the initial calibration phase was cumbersome.
Here, I introduce an improved algorithm which does not necessitate an in-field procedure.
Instead, it achieves camera calibration directly from the recorded data.

%In particular, the neural network has been integrated to my spherifly3D software for the calibration-free 3D reconstruction 

% The trained network has been integrated into a tracking software.
% The software, \verb|oorb|, performs the calibration-free 3D reconstruction of firefly swarms from stereoscopic recordings with 360-degree cameras.

\subsection{Principle of calibration-free calibration}
The details of 360-degree triangulation and calibration-free estimates of camera pose is provided in the Appendix~\ref{appendix}.
Briefly, \verb|oorb| estimates the delay between cameras streams from the cross-correlations of flash time series, and the camera pose (translation and rotation between cameras) by using a calibration set extracted from the movies themselves.

\subsection{Prior results}
An earlier version of the calibration-free algorithm was used to assemble a large dataset of 3D reconstructed flashes and trajectories by crowdsourcing~\citep{sarfati2023crowdsourced}.
Several teams across the United States used pairs of GoPro Max 360-degree cameras, and applied a very simple protocol consisting of placing cameras side-by-side at a set distance, and started recording.
The video were then process to track flashes, and perform calibration from them.
While this proved generally successful, several videos (or parts of the videos) could not be analyzed because they contained too many artifacts.

\subsection{New results}
\begin{figure}
\centering
\includegraphics[scale=0.7]{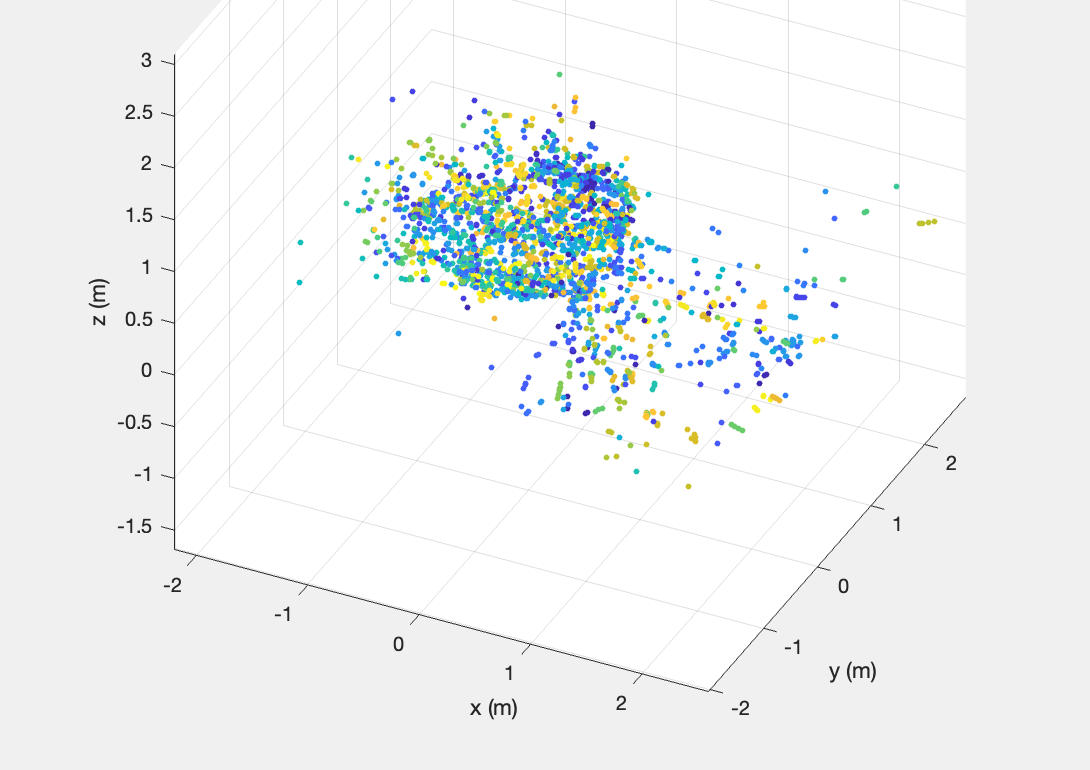}
\caption{
\label{fig:swarm}
3D reconstruction of a firefly swarm (10min recording).
}
\end{figure}

Some videos that could not be processed with the previous pipeline now can be, using \verb|oorb|.
In Figure~\ref{fig:swarm}, I show the 3D reconstruction of a swarm that could not be previously processed.

% \section{Prospects}
% The training set will be expanded to include new data as it comes in.
% In particular, firefly recordings from various iamging sources will be labelled in a close future.
% New trained networks will be provided, ranging from more general to more specialized, for example to focus on photographs.

\section{Conclusion}
% The training set will be expanded to include new data as it comes in; ideally, recordings from various imaging sources will be included.
% New trained networks will be provided, ranging from more general to more specialized, for example to focus on photographs.

Accurately identifying firefly flashes in natural images or movie frames becomes difficult in suboptimal recording conditions.
A neural network trained on a few hundred of manually labeled patches is found to be much more accurate than other thresholding methods.
The training sets and trained networks are provided in order to facilitate flash tracking.
The dataset and code will be maintained and expanded (Python versions are on the way).

Improved firefly tracking will have important implications for firefly monitoring and conservation efforts~\citep{Lewis2020,Fallon2022}.

\section*{Acknowledgments}
Patches were extracted from natural habitat recordings collected notably by Cheryl Mollohan, Ronald Day, and RS, using personal equipment.
I am grateful to Christopher Earls for insightful discussions and support.
There is no funding to report.

\bibliographystyle{iclr2025_conference}
\bibliography{firefleye-net}

\clearpage
\vspace{1cm}
\appendix
\textbf{\LARGE Appendix}

\section*{Methods}
\label{appendix}

%The Methods should include detailed text describing any steps or procedures used in producing the data, including full descriptions of the experimental design, data acquisition assays, and any computational processing (e.g. normalization, image feature extraction). See the detailed section in our submission guidelines for advice on writing a transparent and reproducible methods section. Related methods should be grouped under corresponding subheadings where possible, and methods should be described in enough detail to allow other researchers to interpret and repeat, if required, the full study. Specific data outputs should be explicitly referenced via data citation (see Data Records and Citing Data, below).

%Authors should cite previous descriptions of the methods under use, but ideally the method descriptions should be complete enough for others to understand and reproduce the methods and processing steps without referring to associated publications. There is no limit to the length of the Methods section. Subheadings should not be numbered.

\subsection*{Implementation}
The practical implementation of the method outlined above consists of four major steps: 
1) record fireflies in their natural habitat; %(Fig.~\ref{fig:setup}(a)); 
2) track flashes in each frame of the stereoscopic movies;  %(Fig.~\ref{fig:setup}(b,c)); 
3) triangulate the 3D location of flashes from their complementary planar coordinates;  %(Fig.~\ref{fig:setup}(d)); 
4) trajectorize 3D coordinates by concatenating flashes into trajectories.% (Fig.~\ref{fig:setup}(e)). 
I detail each step below, but first briefly summarize the theoretical foundations of triangulation from spherical movies.

\subsection*{Principles of spherical 3D reconstruction}
The principles and implementation of 3D reconstruction from 360-degree cameras are presented in details in~\citet{Sarfati2020}.
Briefly: 360-degree cameras record at $360^\circ \times 180^\circ$ around them, \textit{i.e.} over the entire sphere. 
The photo-sphere is then rendered as an equirectangular frame of dimension $2P \times P$~pixels$^2$.
The planar coordinates $(w,h)$ map onto the polar~$\theta = 360^\circ \cdot w/2P$ and azimuthal~$\phi = 180^\circ \cdot h/P$ spherical angles. %(Fig.~\ref{fig:setup}bc).
Taking the location and orientation of Camera~1 as the coordinate system of the world, Camera~2 is translated by a vector $\vec{t}$ and rotated by a rotation matrix $\mathbf{R}$ relative to Camera~1.
Let's assume that a flash occurs at a position $\vec{X}_1$ relative to Camera 1. 
Its projection in the field-of-view (FoV) of Camera~1 is $(\theta_1,\phi_1)$, and in the field of Camera~2, $(\theta_2,\phi_2)$.
Based on these projections, $\vec{X}_1$ can be triangulated using the geometric relation:
\begin{equation}
    r_1 \vec{\alpha}_1 - (\vec{t} + \mathbf{R}^{-1}r_2 \vec{\alpha}_2) = 0,
\end{equation}
by solving for $(r_1,r_2)$, with
\begin{equation}
    \vec{\alpha}_i = (\cos\theta_i \sin\phi_i, \sin\theta_i\sin\phi_i,\cos\phi_i).
\end{equation}
Hence, $\vec{X}_1 = r_1 \vec{\alpha_1}$.
Note that we set $| \vec{t} | = 1$, and conversion to real world units requires to rescale by the actual distance between the two cameras.

Matching and triangulating flash positions requires to perform camera calibration, both in time and space.
Specifically, one needs to measure the time delay, in number of frames $\Delta k$, between the two cameras, as well as estimate the camera pose ($\vec{t}$, $\mathbf{R}$).
This calibration was previously done using an artificial flash signal for matching complementary frames, and the trajectory of a small light-emitting diode (LED) for spatial calibration.
These steps required additional experimental steps, and to play the movies manually to extract specific frames and identify the LED trajectory, which was time-consuming and involved significant human input.
This was not suitable for automatic processing of many datasets coming from the large-scale deployment of the stereoscopic setup.

We later found that manual calibration could be avoided by relying instead on the data, \textit{i.e.} recorded firefly flashes, from which $\Delta k$, $\vec{t}$, and $\mathbf{R}$ can all be inferred computationally.
We present this updated method below.

Following automatic calibration and triangulation, flash occurrences can be placed in three-dimensional space, providing in particular density estimations.
From 3D localization, full trajectories, consisting of several flashes from the same individual, can even be constructed~\citep{Sarfati2020}.

\subsection*{Record}
We used GoPro cameras as 360-degree cameras, both the Fusion (older model) and Max (newer model).
After identifying a flat, open location with good firefly activity, the two cameras are placed side-by-side with an approximately parallel orientation (same side facing the same direction). This configuration is recommended since automated calibration by numerical optimization requires a ``guess'' of the relative camera pose to be efficient (see below), in this case $\mathbf{R} \simeq I_3$ (identity matrix).

The two cameras are separated by a distance $s$ and lie a height $h$ above ground. 
We recommend values of $s$ between 1m and 2m, for a good balance of triangulation resolution and field-of-view overlap.
Other separations are acceptable, but in any case the value of $s$ needs to be measured and reported, in order to convert virtual coordinates into real-world units.
Once ready, the cameras are started manually, typically within a few seconds from one another.
The data reported here was captured at 30fps with a high ISO value (typically 6400).

Finally, the recordings are rendered into equirectangular movies (.mp4) using GoPro software.

\subsection*{Track}
Tracking refers to the accurate identification of firefly flashes in each movie frame.
It starts with the detection of clusters of bright pixels
%For reasons presented below, an additional ``cleaning'' procedure is often needed to remove artifacts of various origins.
%\subsubsection*{Flash detection}
%Tracking firefly flashes is a relatively simple procedure.
%It consists of detecting bright pixel blobs on a darker background.
%In ideal recording conditions (perfectly dark environment), it is sufficient to use pixel intensity thresholding.
%However, this method fails in a number of practical situations, notably when recording starts at twilight or in the presence of persistent light pollution (the Moon, artificial lighting such as street or backyard lights, etc.).
%Therefore, we opt for a tracking method that combines 
using adaptive background subtraction and intensity thresholding.
For a given frame $f_k$, a background frame $B_k$ is calculated by averaging all preceding frames within a time $\tau$ from $f_k$ (we used $\tau = 2$s).
The corresponding foreground $F_k$ is then calculated by: $F_k = f_k - B_k$.
Since pixel values are unsigned integers, negative foreground pixels are set at 0. 
A small blurring kernel (typical radius: 1 pixel) is then applied to $F_k$.
Intensity thresholding of the foreground $F_k$ then returns the centroid coordinates of flashes.

\subsection*{Triangulate}

The triangulate procedure encompasses two steps: camera calibration and flash triangulation.
Camera calibration is performed directly from the firefly data (detected flashes). It requires to first estimate the delay between each camera's stream, then estimate camera pose ($\vec{t}$, $\mathbf{R}$).
Flash triangulation requires to first match detected flashes in complementary frames, then triangulate their location from the two sets of planar coordinates.

\paragraph{Temporal calibration.}
Automated frame synchronization relies on the cross-correlation of the time series of the number of detected flashes in each camera, $N_1(k_1), N_2(k_2)$, with $k_i$ the camera time expressed in frame number:
\begin{equation}
    N_1 \star N_2 (\Delta k)= \sum_{k = 0}^{K-k-1} N_1(k+\Delta k) \times N_2(k).
\end{equation}
Assuming first that these two traces are identical but shifted by $\Delta k = k_2-k_1$ frames, the cross-correlation will return the true value for $\Delta k$. %(Fig.~\ref{fig:Tcal}).
%(except in unrealistic situations, \textit{e.g.} if the traces are perfectly periodic).
In practice, however, the time series will be slightly dissimilar as they originate from different FoVs, where visual occlusion and limited light sensitivity impact what each camera records.
With enough data, of the order of $10^5$ frames in a movie, this is not an obstacle.
% The most problematic source of noise comes from occasional background light pollution.
% Typically, this will occur at the beginning of the recordings, if the night is not yet obscure enough or if a flashlight is used to set up the cameras. 
% (The latter might happen at the end as well.)
% Additionally, there are sometimes persistent objects in the FoV background, \textit{e.g.} Moon, backyard light, passer-by with a flashlight, etc.

% To alleviate these issues, we start by pre-processing the original input, consisting of a set of $(w,h,k)$ coordinates.
% %, with $(w,h)$ the flash coordinates in the equirectangular frame (with, height).
% The sets are truncated to remove the points before the first $N=0$ and after the last $N=0$ frames, which eliminates sky brightness and flashlight effects (Fig.~2c).
% Secondly, a simple tracking algorithm is applied to detect persistent objects lasting over a certain duration which would be impossible for a firefly to sustain, and therefore corresponds to light pollution (Fig.~2d).
% The threshold is set to 300 frames (5s or 10s, depending on frame rate), so as not to exclude glowing species, such as ``blue ghost'' fireflies (\textit{Phausis reticulata}).

To make sure that the time delay estimation remains robust against possible residual noise, and estimate the reliability of the measurement, we apply a Random Sample Consensus-type algorithm (RANSAC) to the cross-correlation procedure: 100 random intervals are extracted from the cleaned traces, and cross-correlation is applied to every pair.
The modal value becomes the $\Delta k$ used to align the frame streams.

\paragraph{Spatial calibration.}
Spatial calibration requires a set of matched points\footnote{Ideally, spatially dispersed.} 
so that $(\vec{t},\mathbf{R})$ can be estimated according to the method described in~\citet{Sarfati2020}.
Originally, matched points were obtained from an identifiable trajectory, created manually with a small LED at the beginning of the recording. %(Fig.~\ref{fig:Scal}).
For the automated calibration algorithm, a set of matched points is collected simply by extracting all positions where $N_1(k_1)=N_2(k_2-\Delta k)=1$ in complementary frames. %(Fig.~\ref{fig:Scal}).
Unique flashes ($N = 1$) are very common as fireflies flash intermittently and at low densities.
We usually limit the calibration set to 1000 pairs of points for faster processing.
Camera calibration is then achieved by numerical optimization of a cost function, as explained in~\citet{Sarfati2020}.

% Most of the time, these unique flashes will indeed correspond to the same firefly, and will be a true matched point.
% Sometimes, however, they will come from two different sources.
% Consequently, the algorithm must be robust against outliers.
% For this reason, we employ a Maximum Likelihood Estimator for $(\vec{t},\mathbf{R})$. 
% Taking as input a set of points, the algorithm calculates the values for $(\vec{t},\mathbf{R})$ that maximize a likelihood function.
% As a global optimizer, outliers contribute little to the overall cost.

\paragraph{Matching and triangulation}
The matching and triangulation algorithm is identical to the one explained in~\citet{Sarfati2020}.

\subsection*{Trajectorize}

\paragraph{Flashes to streaks}
I define as \textit{streak} the spatiotemporal span of a firefly flash as it appears in consecutive frames. 
For two flash coordinates to belong to the same streak, they must appear in consecutive frames and be separated by less than a distance threshold~$d_\mathrm{max}$ (typically 0.3m).
%set at 0.3m.
This is readily achieved by computing a connectivity matrix $\mathbf{C}$ from the pairwise delays and distances: 
\begin{equation}
C_{ij} = (\Delta t_{ij} == 1\mathrm{frame}) \, \mathrm{AND} \, (\Delta r_{ij} < d_\mathrm{max}).
\end{equation}

\paragraph{Streaks to trajectories}
Some species of fireflies typically produce several flashes in a row (`flash trains'), rather than single flashes. 
I define as trajectories the spatiotemporal span of flash trains emitted by a single individual.

The concatenation of streaks into trajectories relies on distance-based linkage: streaks within both a certain time span $\Delta t$ and a certain distance $\Delta r$ from one another are considered part of the same trajectory. 
Generally, I use $\Delta t = 1$s and $\Delta r = 
1$m.

This is the most versatile approach. 
It works well because fireflies are generally in sufficiently low density that ambiguities are rare.
More sophisticated approaches are possible, for example by including kinematics variables such as speed and acceleration, as presented in~\citet{Sarfati2020}.

% \subsection*{Subsection}

% Example text under a subsection. Bulleted lists may be used where appropriate, e.g.

% \begin{itemize}
% \item First item
% \item Second item
% \end{itemize}

% \subsubsection*{Third-level section}
 
% Topical subheadings are allowed.

% \subsection*{Crowdsourcing protocol}

% Data contributors unaffiliated with the University of Colorado were given simple instructions regarding the placement and operation of cameras in the natural environment.
% Briefly, they were instructed to: 1) identify a suitable site within the firefly habitat; 2) place the GoPro cameras side-by-side and parallel to one another; 3) measure and report the distance between them and their height above ground; 4) start recording once flashes started.
% The exact protocol provided (for 2022) is available in the dataset.

% Most contributors were lent a pair of GoPro cameras along with an external hard drive to offload and store recordings. 
% They were given the liberty to choose the dates and locations of their recordings based on their knowledge and interests.
% They were asked to report these specifics, and if possible other contextual information such as weather conditions or identification of firefly species.

\end{document}